\documentclass[10pt,twocolumn,letterpaper]{article}

\usepackage{wacv}
\usepackage{times}
\usepackage{epsfig}
\usepackage{graphicx}
\usepackage{amsmath}
\usepackage{amssymb}
\usepackage{tabularx}
\usepackage{multirow}
\usepackage{xcolor}
\usepackage{listings}
\def\wacvPaperID{1837} % Enter the WACV Paper ID here

\wacvfinalcopy
\ifwacvfinal
\def\assignedStartPage{1} % *** Enter the assigned starting page number (instead of 9876)
\fi
\ifwacvfinal
\usepackage[breaklinks=true,bookmarks=false]{hyperref}
\else
\usepackage[pagebackref=true,breaklinks=true,colorlinks,bookmarks=false]{hyperref}
\fi

\ifwacvfinal
\else
\fi

\pdfoutput=1

\begin{document}

\title{Texture Extraction Methods Based Ensembling Framework for Improved Classification}

\author{Vijay Pandey\\
IBM\\
India\\
{\tt\small vijay.pandey1@ibm.com}
\and
Trapti Kalra\\
IBM\\
India\\
{\tt\small trapti.kalra@ibm.com}
\and
Mayank Gubba\\
NIIT University\\
India\\
{\tt\small Mayank.gubba18@st.niituniversity.in}
\and
Mohammed Faisal\\
NIIT University\\
India\\
{\tt\small mohammed.faisal18@st.niituniversity.in}}
\maketitle

%%%%%%%%% ABSTRACT
\begin{abstract}

Texture-based classification solutions have proven their significance in many domains, from industrial inspections to health-related applications. New methods have been developed based on texture feature learning and CNN-based architectures to address computer vision use cases for images with rich texture-based features. In recent years, architectures solving texture-based classification problems and demonstrating state-of-the-art results have emerged. Yet, one limitation of these approaches is that they cannot claim to be suitable for all types of image texture patterns. Each technique has an advantage for a specific texture type only. To address this shortcoming, we propose a framework that combines more than one texture-based techniques together, uniquely, with a CNN backbone to extract the most relevant texture features. This enables the model to be trained in a self-selective manner and produce improved results over current published benchmarks- with almost same number of model parameters. Our proposed framework works well on most texture types simultaneously and allows flexibility for additional texture-based methods to be accommodated to achieve better results than existing architectures. In this work, firstly, we present an analysis on the relative importance of existing techniques when used alone and in combination with other TE methods on benchmark datasets. Secondly, we show that Global Average Pooling- which represents the spatial information- is of less significance in comparison to the TE method(s) applied in the network while training for texture-based classification tasks. Finally, we present state-of-the-art results for several texture-based benchmark datasets by combining three existing texture-based techniques using our proposed framework.
\end{abstract}

\section{Introduction}
Image texture analysis plays a vital role in multiple use cases. It has been used in medical imaging \cite{liu2019bow}, remote sensing \cite{zhai2020deep}, industrial inspection of materials \cite{silven2000applying}, autonomous vehicles \cite{zu2015real}, explosive hazard detection \cite{petersson2016multi} and other industries. Image texture has numerous definitions which defines unique ways to calculate discrete qualities of texture like smoothness, roughness, uniformity, homogeneity, etc. These methods are based on the concept of repeating patterns in an image \cite{liu2022deep, tuceryan1993texture}. One use of texture analysis is that the Image's texture classification enables better Image classification. The spatial distribution of pixels from an image is an integral part of texture classification \cite{julesz1981textons}. Convolutional Neural Network (CNN) has been very successful in capturing the local and global spatial features, which play a key role in texture classification. CNN preserves the relative spatial information with the help of convolution layers and aggregates the spatial information using the pooling layers \cite{xue2018deep}.

Local spatial features assist in recognizing the patterns of a texture. In an image, while classifying the texture, some local patterns are repeated throughout the image \cite{tuceryan1993texture}. These local features exhibit almost the same characteristics as one another and they are extremely critical in differentiating the textures. The traditional CNN architectures do not exhibit phenomenal results for classifying the textures merely based on the spatial features.

The CNN can separate the local features with the aid of convolution layers although these spatial features are aggregated by employing the pooling layer. They make decisions primarily based on the global features rather than focusing on the local features. CNN tends to lose the locally rich features and that’s the major reason behind the under-performance of CNN while identifying textures \cite{chen2021deep}. To overcome this problem, numerous techniques have been introduced where texture extraction (TE) layers are applied before the fully connected (FC) layers, using which local features are extracted. They are used along with global features. Several techniques such as DEPNet \cite{xue2018deep}, DeepTEN \cite{zhang2017deep}, CLASSNet \cite{chen2021deep}, FENet \cite{xu2021encoding}, etc., have been built based on this strategy.

Each of these methodologies focuses on some crucial aspects of texture and is specialized for specific use-cases. Using a Histogram layer \cite{peeples2021histogram} along with CNN produces adequate results, whereas FENet \cite{xu2021encoding} produces the state-of-the-art (SOTA) results for some benchmark texture datasets. In this paper, inspired by DEPNet \cite{xue2018deep} where the the novelty is an ensemble of GAP and DeepTEN \cite{zhang2017deep} techniques with bi-linear pooling delivering superior results, we propose a method where focused on combining several techniques rather than developing a new one. When various TE processes are combined, we can effortlessly capture the divergent characteristics of a texture in a better manner as each of these processes emphasizes unique aspects of texture. We achieved SOTA results on the major texture datasets by integrating distinct existing approaches i.e., DeepTEN, Histogram, and FAP \cite{zhang2017deep, peeples2021histogram, xu2021encoding}. These three techniques have been used with Global Average Pooling (GAP) in their original papers respectively. In our work, we have ensembled these techniques while removing GAP from each technique and tuning the size of their activation and hyper-parameter values to produce an optimized network with better generalization. By the implementation of this strategy, we achieve superiority over each method individually, since we enable classification of texture with distinct characteristics rather than an architecture focusing on any one major trait of texture.
\section{Contribution}
% In machine learning, it is well known that a group of distinct and unique weak learners when ensembled together may form a strong learner. 
In this work we suggest a framework where more than one TE technique has been ensembled to achieve state-of-the-art (SOTA) results on bench-marked texture-based datasets.

In prior available research work generally two TE methods have been used in an ensembling architecture. However, we propose a framework that enables ensembling more than two techniques at the same time. For feature aggregation, prior research uses BLP but we apply concatenation- in view of the fact that it reduces the number of parameters in the network. GAP has very less impact on the model when it is concatenated with another texture feature extraction technique. Whilst we are concatenating distinct techniques post convolution layer, we can replace the GAP with some other mechanism considering it makes a minimal impact on the model's accuracy. Finally, using the proposed framework we have ensembled numerous feature extraction techniques demonstrating SOTA results on text-based bench-marked datasets.

Our proposed framework fixes a known design challenge. When an AI expert needs to solve a texture feature extraction problem or improve the results for a texture based dataset model, the first approach is typically to hand craft extraction of texture patterns, which is complex and requires domain expertise, involves experimentation to evaluate the diverse text extraction techniques. Our technique removes the need for hand crafting the extraction techniques by providing a holistic architectural solution which can extract a wide variety of textures in a diversified texture dataset.

We also demonstrate that unlike the other cited papers, GAP is not significant for texture specific images by using multiple experiments, which is a new theory on significance of GAP for texture-based images.
\section{Main Idea}
There is no single perfect method that works for all texture-based datasets. Combining multiple techniques specialized in learning different texture representations from images can enable improved learning. Inspired by the idea of passing related inputs along with the original input to yield better results, Pandey et al. \cite{pandey2020incorporating}, we employ the idea to pass additional informative features to enable better learning. By ensembling varied TE techniques, applied on top of the backbone architecture we build a model which is more efficient. This ensemble works by classifying texture based on a variety of texture features which are acquired using the individual TE techniques. Our experiments demonstrate that merging features extracted from dissimilar techniques improves the performance of model in comparison to an individual TE methods applied in isolation. Texture is the key feature for image classifications, and texture-based classification can assist in image classification tasks in many applications. Given the wide range of texture types embedded in different image datasets, we aim to construct a flexible and scalable model which can be used for a variety of texture-based datasets. From the previously developed techniques namely DeepTEN: Texture Encoding Network \cite{zhang2017deep}, DEPNet \cite{xue2018deep}, DSRNet \cite{zhai2020deep}, etc. We can say that applying the texture feature extraction techniques after the activation layer of a backbone results in a better accuracy compared to techniques where the backbone is not used. Once the input image is passed through the backbone i.e., pre-trained weights of the convolution and pooling layers of the backbone model, subsequently local features, as well as the global features, can be examined to extract texture efficiently to classify the texture with improved preciseness. To produce an efficient ensemble model, we need to choose our methods astutely, as we don’t want to combine two techniques that are competent at doing the same task. We need to select the TE techniques in such a way that there is less correlation between the techniques to maximize the utilization of each technique while providing the best possible output. According to our idea, each technique which is chosen for the ensembled model should work on separate aspects of the data. This will ensure the smooth working of the model and will ensure that the resultant model is better than individual models. Using any specific technique for TE limits its use-case for a specific variety of textures. For other texture types that technique may not work. If one can use various methods in a single architecture then it would be able to handle complex texture types.
\section{Related Work}
Global Average Pooling (GAP) has been used for feature aggregation- TCNN \cite{andrearczyk2016using} and wavelet CNN \cite{fujieda2017wavelet} are the most well known applications of GAP. This method is lossy regarding spatial information but is suitable for extracting the texture features \cite{fujieda2017wavelet}. The bi-linear pooling (BLP) introduced by Lin et al. \cite{lin2016visualizing} makes second-order interactions between the two outputs \cite{dai2017fason}- it is an outer product of each pair of feature points. BLP encodes more information than GAP. GAP also acts as a structural regularizer which enables us to explicitly enforce feature maps to classify the textures more efficiently \cite{lin2013network}. DeepTEN \cite{zhang2017deep} is a technique that uses residual dictionary learning to encode features on an end-to-end learning framework. It is different from SIFT \cite{lowe2004distinctive} and filter banks \cite{cimpoi2015deep} as hand-engineered features are used, but in DeepTEN method the dictionaries, features, classifiers, and encoding representation are all trained together to produce an orderless encoding. DeepTEN is very efficient for material and texture recognition tasks, and can also be used as a pooling layer on top of convolution layers to increase the flexibility of the model \cite{liu2019bow}. To integrate histograms with CNN, a histogram layer was introduced by Joshua \cite{peeples2021histogram}, which captures the local spatial features using a histogram. The histogram layer uses a Radial basis function (RBF) \cite{bors2001introduction} for the binning operation to find the bin centers and widths. It also makes use of the GAP to capture the spatial, texture, and orderless convolution features. The histogram layer enables the model to use fewer layers than conventional CNN, as the texture information is directly captured by feature maps \cite{peeples2021histogram}. Fractal analysis pooling (FAP) is used in FENet to group points of a texture based on the local fractal dimension of the image. It eliminates the dependency on spatial order while encoding the characteristics of features \cite{xu2021encoding}. Fractal dimension is a quantitative measure of the roughness of a given image \cite{shanmugavadivu2012fractal}. GAP is used to capture the spatial features of a texture but sometimes it fails to distinguish between complex distributive patterns from a texture. FENet uses both GAP and FAP as the method and distributive features are captured from distinct aspects \cite{xu2021encoding}. Similarly, in this work, we propose a model architecture which exhibits an improvement in results by combining several techniques after the convolution layers. In the proposed model we consider the TE methods which extract properties from a single feature map unlike the class-net \cite{chen2021deep} where numerous feature maps are used to find the texture features.

\section{Ensemble of TE Techniques}
\subsection{Mathematical Formulation}
With input $X$ provided to the backbone $B$ in the network,
\begin{align}
    X_{B} = B(X)
\end{align}
$X_B$ is the activation output of the last layer of the backbone. As $X_B$ contains the common low level features of the image, to achieve improved results on the texture classification task, extracting texture features from $X_B$ and passing it to FC layers will help. Given that we are using $K$ different methods for TE. Each method is denoted as $M_i$ and output of any method $M_i$ will be written as,
\begin{align}
    \label{eq:mi}
    X_{Mi} = M_i(X_B), \forall i \in [1, K]
\end{align}
There is an aggregation function $A$, which aggregates the output obtained from $K$ TE methods. Output of aggregation function can be formulated as,
\begin{align}
    X_A = A(X_{M1}, X_{M2}, \dots, X_{Mi}, \dots, X_{Mk})
\end{align}
$X_A$ will be passed as an input to the FC layer. In our paper $A$ is the concatenation function $CA$ which can be defined as,
\begin{align}
CA(X_{1}, X_{2}, \dots, X_{i}, \dots, X_{k}) = X_{1}X_{2}\dots X_{i}\dots X_{k}
\end{align}
Size of $X_{CA}$ can be defined as,
\begin{align}
\label{eq:ca}
|X_{CA}| = |X_{M1}| + \dots + |X_{Mi}| + \dots + |X_{Mk}|
\end{align}
\textbf{Bilinear models} The factors in bilinear models \cite{freeman1997learning} keep the contributions of the two components balanced.
Let $a^t$ and $b^s$ represent texture information of surface and spatial information with vectors of parameters and with dimensionality $I$ and $J$. Then the bilinear function $Y^{ts}$ is,
\begin{align}
Y^{ts} = \displaystyle\sum_{i=1}^I \displaystyle\sum_{j=1}^J w_{ij}a_{i}^{t}b_{j}^{s}
\end{align}
where $w_{ij}$ is a learnable weight which balances the interaction between surface texture and spatial information. The outer product notion captures a pairwise correlation between the surface texture encodings and spatial observation structures. Suppose $A$ is the bilinear model $BM$ then,
\begin{align}
\label{eq:bm}
|X_{BM}| = |X_{M1}|* |X_{M2}|
\end{align}
\subsection{Ensembling Texture Extraction Techniques}
\begin{figure}[t]
\begin{center}
\includegraphics[scale=0.3]{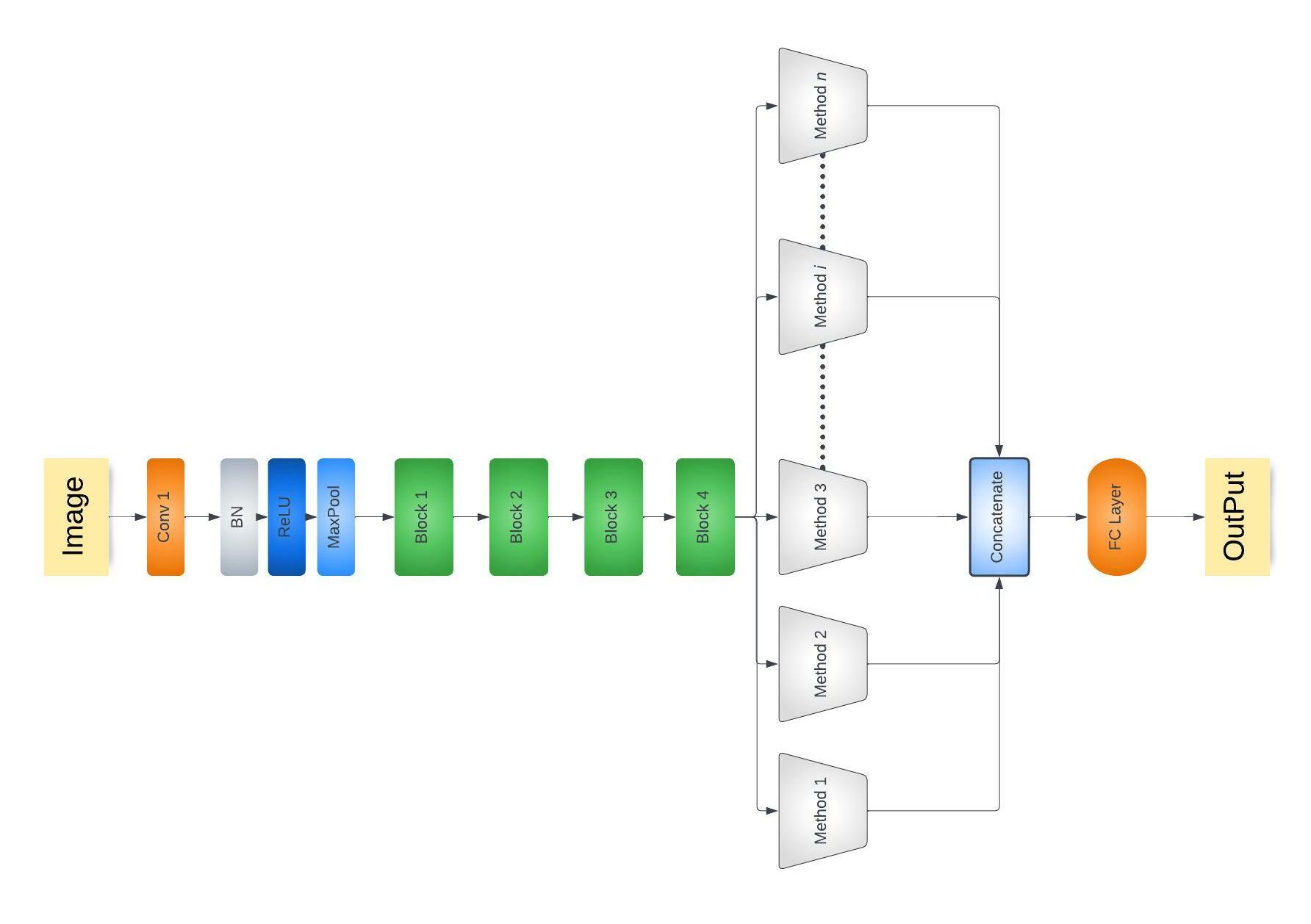}
\end{center}
\caption{Framework for combining more than one technique}
\label{fig:framework}
\end{figure}

We propose an ensemble framework in which the selected TE methods should be unique in their working principle and should not have overlapping mechanisms. Failing on this aspect will result in less accuracy because of method similarity which provides overlapping features. 
\begin{figure}[t]
\begin{center}
\includegraphics[scale=0.7]{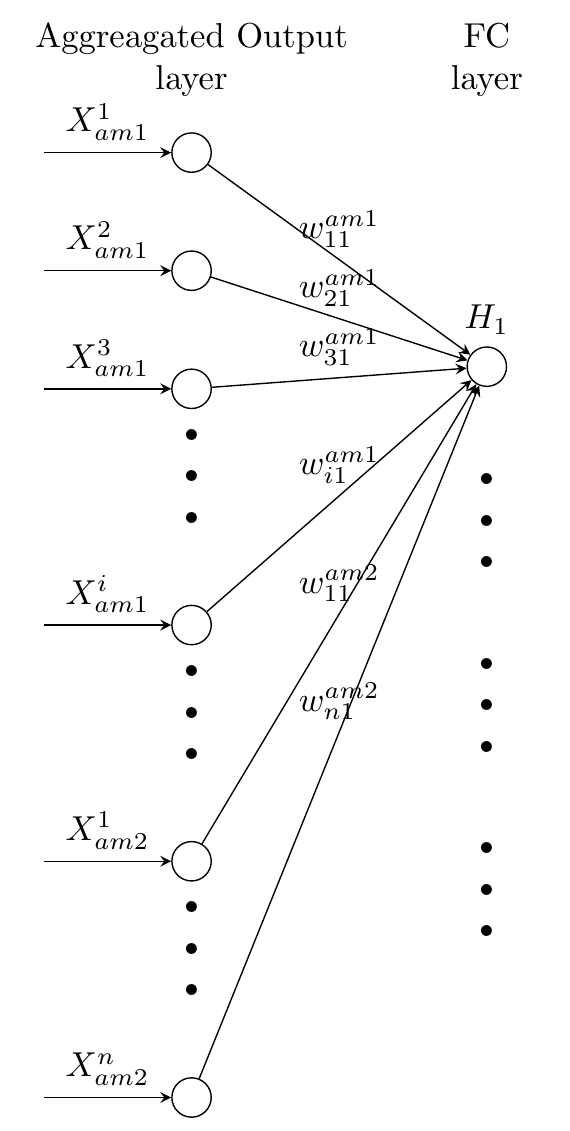}
\end{center}
\caption{Illustration of interaction between aggregated output and FC layer}
\label{fig:A_FC}
\end{figure}
%here we show the diverse technique principle
To understand how using diverse techniques helps in improving accuracy, follow the Figure \ref{fig:A_FC}. If we formulate this figure mathematically the equation becomes (ignoring bias term),
\begin{align*}
    Y_{H1} = X_{am1}^{1}*w_{11}^{am1} + X_{am2}^{1}*w_{11}^{am2} + X_{am1}^{2}*w_{21}^{am1} +\\ X_{am1}^{3}*w_{31}^{am1} + X_{am1}^{i}*w_{i1}^{am1} +  X_{am2}^{n}*w_{n1}^{am2}
\end{align*}
The above equation looks similar to a regression problem- as formulated on the single hidden layer of the FC layer. To get better weights in this equation, correlation between the features should be as less as possible. In our case, to achieve less correlation, activation output from different methods should be less similar. It shows that diverse methods capturing distinct type of textures enable model to give better results.
%here we detail about the framework
Feature extraction techniques are used after the convolution layers of backbone network, as shown in Figure \ref{fig:framework}. After passing the input to the backbone network, the output from the final block is passed as an input to these unique TE methods. Several feature extraction strategies can be used as long as they are unique and distinct in extracting texture features. Before passing the outputs of the feature extractor to FC layers, the outputs from multiple texture feature extraction layers are concatenated. Lastly, the output from the FC layer is passed to a softmax layer to  predict the output.

\begin{figure}[t]
\begin{center}
\includegraphics[scale=0.3]{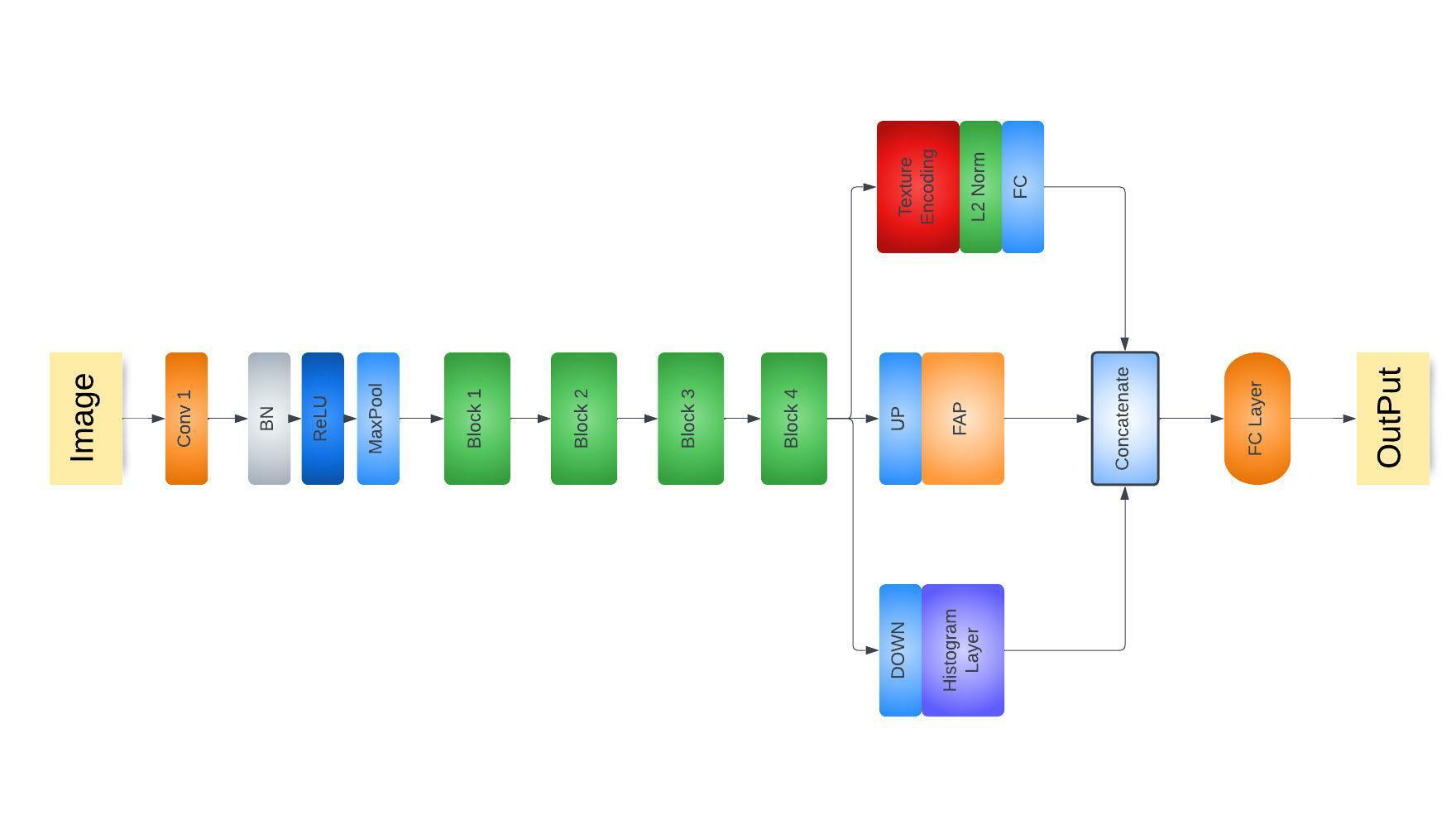}
\end{center}
\caption{Ours proposed architecture}
\label{fig:proposed_architecture}
\end{figure}

The framework discussed in Figure \ref{fig:framework} is used as a model to produce SOTA results on known benchmark texture datasets. Three unique TE techniques are ensembled— DeepTEN \cite{zhang2017deep}, FAP \cite{xu2021encoding}, and Histogram Layer \cite{peeples2021histogram}. We are able to produce SOTA results for different datasets which will be discussed in the Results section. In further experimentation, other combinations of TE methods are considered, and based on the results it is observed that when GAP is not used, the accuracy of the model tends to increase.
%here we described our architecture more specifically
The architecture of our model is described in Figure \ref{fig:proposed_architecture} where output from the last block of backbone is passed to three separate feature extractors. 1) Downsampling layer followed by histogram layer, 2) Feature extractor is upsampled and passed to a fractal analysis pooling (FAP) layer, 3) Texture encoding layer passed to a normalization layer followed by FC layer. The upsampling and downsampling are done to match the input size between block 4 and FAP, and between block4 and histogram layer respectively.

In our architecture we have not included GAP technique unlike the most of the contemporary architectures. As seen in Table \ref{tab:gapvsours}, when GAP is precluded while ensembling the other three techniques the accuracy increases with good percentage on FMD and KTH datasets on the contrary when GAP is combined with other three techniques then accuracy decreases. In the earlier methodologies, GAP is used to add the spatial information, however our research shows that for texture based networks GAP is not of much significance compared to the TE methods. As GAP captures the global spatial information of an image which does not rely on the local features. However, texture patterns are classified based on the local features. Thus, GAP is not of much relevance in texture classification tasks. This shows that while ensembling feature extractors, the the methods combination must be critically analyzed. Though individual methods may produce good results, the combination of unique techniques may offer a scope of improvement.
\begin{table}
\begin{center}
\begin{tabular}{|c|c|c|} %to specify number of columns
\hline %for horizontal line after row
\textbf{Dataset} & \textbf{With GAP} & \textbf{Without GAP } \\ % specifying the column values separated by &
\hline
\textbf{FMD \cite{sharan2009material}} & 82.0 & 83.1\\
\hline
\textbf{KTH \cite{caputo2005class}} & 86.7 & 88.04\\
\hline
\end{tabular}
\end{center}
\caption{Comparison between techniques with and without GAP}
\label{tab:gapvsours}
\end{table}

\subsection{Bi-linear Pooling v/s Concatenation}
BLP is utilized as the second-order interactions between two inputs. One drawback of using second-order interaction is it can accept only two inputs to participate, which limits us from combining more than two inputs. Secondly, provided that the same inputs are passed to BLP and concatenation, the output size of BLP is more when compared to concatenation operation. 
Suppose that, value of $|X_{Mi}|$ as per \ref{eq:mi} is $n$ $\forall i \in [1, K]$, where $K$ is total number of methods used. Comparing equation \ref{eq:ca} and \ref{eq:bm} we get, $|X_CA| = 2*m$ and $|X_{BM}| = m^2$ for $K=2$. We can see that using concatenation aggregation function results in a linear increase in the size of activation as the number of methods increase, on the contrary applying bilinear pooling as an aggregation function results in a polynomial increase in the activation size as the number of methods increase. Due to the bigger output size of BLP, the model becomes large and thus increases the number of parameters. Thus, while combining two TE techniques the output size in BLP will be greater than or equal to concatenation i.e., $m*n\geq m+n$. Secondly, concatenation method can combine more than two techniques at the same time. We can apply BLP for more than two inputs, but it would not be explainable to make second-order interaction between two inputs while leaving other inputs as it is. Thus we prefer to use concatenation rather than BLP in our proposed framework. BLP is not a suitable method when more than two TE techniques are used to extract features.
\subsection{TE Techniques Selection}
While ensembling the TE techniques diversity of methods is the key to enable optimal feature extraction across and variety of input types. Also the distinct methods should
be compatible with the same backbone architecture. During the selection of different TE techniques, we need to ensure that the difference between the size of activation outputs is not large, as it would impact the performance of the technique having smaller size activation output. For instance, combining two techniques $A$ with an activation output of size of 1024 and
$B$ with an activation output size of 32 will exert more importance on technique $A$ and thus dominating method $B$. Pointwise convolution technique can be used to reduce the activation output size to make it compatible to use it with other techniques.
\subsection{Selected Techniques and Hyperparameters}
\textbf{Histogram}
Used the implementation as per the paper \cite{peeples2021histogram}. Output is taken from the last block of ResNet18. As per the paper \emph{dim} is 2 for image data, \emph{number of bins} is 4, \emph{bin width} is 128 for ResNet18. The input size for the convolution layer is calculated by the formula, $$ \frac{feature\_maps}{feature\_map\_size * number\_of\_bins} $$ which would be 512/(4*4) = 32 for ResNet18. The output size of the convolution layer is calculated as a product of the number of bins and the input size. In our case it is 4*32=128 for ResNet18. For compatibility of the model grouping is important, we set it as 512 groups.
\newline
\newline
\textbf{DeepTEN}
Applied the DeepTEN technique as per the paper \cite{zhang2017deep}. The encoding layer accepts features or feature channels of dimension 512. The number of codewords ($n\_codes$) used in our case is 8. Next is the linear layer with input features as 512*$n\_codes$ which is 4096 and output features as 128. Lastly, batch normalization is done with 128 features.
\newline
\newline
\textbf{Fractal}
Used FAP module as per the hyperparameter settings and config in paper \cite{xu2021encoding}, with \emph{dim} value as 16.
\newline
\newline
\textbf{GAP}
Global Average Pooling (GAP) is a pooling operation designed to capture spatial information of an image in classical CNN. We applied a 2$D$ average pool over an input signal composed of several input planes with the kernel size 7. Added a linear layer to apply a linear transformation to the incoming data with input features as 512 and output features as 48 followed by batch normalization.

\subsection{Reasons to include Four Discussed Techniques }
We have chosen DeepTEN \cite{zhang2017deep}, histogram layer \cite{peeples2021histogram}, FAP \cite{xu2021encoding}, and GAP TE techniques for the proposed model design and experiments. Other methods such as DEPNet \cite{xue2018deep}, DSRNet \cite{zhai2020deep} and CLASSNet \cite{chen2021deep} are inherited from one of these used methods. Our choice of TE methods prevents any overlap in capturing the texture characteristics. The other existing methods leverage outputs from multi layers of the backbone network simultaneously; while our framework focuses on techniques applied to the backbone's last layer output. Some techniques are not included due to similarity with above techniques. DEPNet and DSRNet use the same encoding layer that has been used in DeepTEN. DSRNet \cite{xue2018deep} which was developed for ground terrain recognition uses the same texture encoding layer from DeepTEN. DSRNet has been built using ResNet50 \cite{he2016deep} as a backbone but not uses ResNet18, but in our paper we have used ResNet18 \cite{zhai2020deep}. Whereas, CLASSNet uses a concept of statistical self-similarity (SSS) \cite{mandelbrot1967long} which is calculated with the help of fractals which is something similar to FAP. In CLASSNet the feature maps are learned using CNN and have a cross-layer SSS for exploiting the SSS of images. For the SSS-based texture recognition, fractal analysis is used to characterize the SSS on images \cite{chen2021deep}. The fractal analysis is also used in FAP for calculating the fractal dimension and classifying the texture based on the same.

Considering the other methods as just an extension of the already used methods or not used in the ResNet18 as backbone, we proceed to use the parent methods since we are ensembling the distinct techniques and we don’t want the texture to be categorized based on the same attributes more than once.

\section{Comparison with Other Texture Extraction Techniques}
Our proposed model concentrates on ensembling texture extractors with less correlation among them. We capture the unique texture characteristics of an image useful for classifying the images per their labels. GAP is a pooling layer and has the capability to be used in many techniques \cite{fujieda2017wavelet, peeples2021histogram, xu2021encoding} to capture local-to-global spatial features, however as demonstrated it in not useful in TE techniques. The drawbacks of BLP v/s Concatenation are mentioned in section 5.3. With concatenation, we can preserve information from the previous layer in the same form. When there is minimal correlation between the characteristics extracted we can not afford to lose any information extracted by each of these techniques. For this reason concatenation is used to combine the DeepTen \cite{zhang2017deep}, Histogram layer \cite{peeples2021histogram} and FAP \cite{xu2021encoding} to produce SOTA results on benchmark datasets such as KTH \cite{caputo2005class}, FMD \cite{sharan2009material}, and MINC.

In DeepTEN feature extraction, dictionary learning, and encoding representation training occurs simultaneously. The learned convolution features are easily transferred since the encoding layer learns an inherent dictionary that carries domain-specific information. DeepTEN is a flexible framework that allows an arbitrary input image size making it easier to combine with any model. FAP focuses on discriminating textures based on the fractal dimension i.e., based on the roughness of the texture. As mentioned earlier, histogram layer captures the texture information directly from the feature maps and is based on the fundamentals of local histograms which can be used to distinguish textures. All these techniques emphasize different texture characteristics for classification tasks thus ensembling these unique techniques enables elegant classification. When various characteristics are taken into consideration then our model has more rich information which helps to draw a more generalized boundary between distinct texture classes.
\section{Experiment Evaluation}
\subsection{Experiment Setup}
To test the effectiveness of our model we conduct the experiments on 6 benchmark datasets KTH \cite{caputo2005class}, FMD \cite{sharan2009material}, DTD \cite{cimpoi2014describing}, MINC \cite{bell2015material}, GTOS \cite{zhang2019learning}, and GTOS-M \cite{xue2018deep}. For FMD, DTD, MINC, GTOS, and GTOS-M we use the recommended data split plan. For KTH we generate 10 random splits in the ratio of 3:1 for train:test. The mean and standard deviation for the results are calculated for all the splits and presented in the table  \ref{tab:SOTA}.
ResNet-18 is used as a backbone and the model is trained using cross-entropy loss via an SGD optimizer and cosine scheduler for KTH, DTD, and GTOS. CosineAnnealing warm restart \cite{loshchilov2016sgdr} scheduler is used for FMD, MINC, and GTOS-M. The model is trained for 30 epochs for DTD, FMD and KTH while for MINC, GTOS and GTOS-M the model is trained for 20 epochs. We use a batch size of 32 for KTH, 16 for FMD, 64 for DTD, MINC \& GTOS and 128 for GTOS-M. Learning rate has been set to $1e^{- 3}$ on FMD, $1e^{-2}$ on DTD, $5e^{-2}$ on GTOS-M and $5e^{-3}$ on KTH, MINC and GTOS. As part of the data augmentation the training dataset is transformed by resizing images into 256, random horizontal flipping, random cropping (five cropping technique for DTD and GTOS) of size 224, and finally by normalizing the images.
The five crop augmentation method works well as texture is a neighborhood pixels based feature- cropping enriches the training data. It does not exhibit good results on MINC dataset because some shape based features are also present along texture based in this dataset. The model is built using Pytorch 1.7.1 version and the experiments presented here were run on NVIDIA DGX system with 8 x A100 32 GB GPUs.
\subsection{Results}
On comparing our proposed model with well-known texture-based classification models CLASSNet \cite{chen2021deep}, DeepTEN \cite{zhang2017deep}, DSRNet \cite{xue2018deep}, MAPNet \cite{zhai2019deep}, LSCNet \cite{bu2019deep}, DEPNet \cite{xue2018deep}, HistNet \cite{peeples2021histogram}, and FENet, the results obtained are presented in Table \ref{tab:SOTA}). The results for TE methods used for comparison have been quoted directly from the respective papers and left blank where corresponding results are not available.
Our method gives SOTA accuracy for FMD, KTH, and MINC for the ResNet18 backbone. Standard deviation for KTH is higher due to the randomness in the splitting of
the data. In comparison to the other methods, our model increases the accuracy on benchmark datasets by a significant percentage. We also produce notable results for DTD, GTOS and GTOS-M datasets- our model yields competitive results against the current SOTA benchmarks. For GTOS our model is only behind the CLASSNet reported accuracy. Apart from accuracy metric, our architecture is more efficient compared to the current models in terms of the training time. Currently our proposed model takes less time to train large datasets such as MINC, GTOS, GTOS-M producing the model's best accuracy within 20 epochs of training with similar batch sizes as used in published papers.
\begin{table}[t!]
\begin{center}
\scalebox{0.8}{%
\begin{tabular}{|l|l|l|l|l|l|}
\hline
\textbf{Dataset} & \textbf{DeepTEN} & \textbf{GAP} & \textbf{Histogram} & \textbf{FAP} & \textbf{Accuracy} \\ \hline
\multicolumn{1}{|c|}{} & {yes} & {Yes}  & {Yes} & {Yes} & 82.0 \\
\multicolumn{1}{|c|}{} & {No} & {Yes}  & {Yes} & {Yes} & 80.7 \\
% \cline{2–6}
\multicolumn{1}{|c|}{} & {Yes} & {No} & {Yes} & {Yes} & \textbf{83.1} \\
% \cline{2–6}
\multicolumn{1}{|c|}{} & {Yes} & {Yes} & {No}& {Yes} & 81.9 \\
% \cline{2–6}
\multicolumn{1}{|c|}{\multirow{-4}{*}{FMD}} & {Yes} & {Yes} & {Yes} & {No} & 82.2 \\ \hline
\multicolumn{1}{|c|}{} & {yes}& {Yes} & {Yes} & {Yes} & 86.7 \\
\multicolumn{1}{|c|}{} & {No}& {Yes} & {Yes} & {Yes} & 86.4 \\
% \cline{2–6}
\multicolumn{1}{|c|}{} & {Yes} & {No}& {Yes} & {Yes} &  \textbf{88.04} \\
% \cline{2–6}
\multicolumn{1}{|c|}{} & {Yes} & {Yes} & {No} & {Yes} & 87.1\\
% \cline{2–6}
 {\hfil\multirow{-4}{*}{KTH} }& {Yes} & {Yes} & {Yes} & {No} & 87.8\\ \hline
\end{tabular}}
\end{center}
\caption{Comparing ensemble of three out of four techniques at once}
\label{tab:3_by_4}
\end{table}
\begin{table}
\begin{center}
\scalebox{0.8}{%
\begin{tabular}{|l|l|l|l|l|l|}
\hline
\multicolumn{1}{|l|}{\textbf{Dataset}} &
\textbf{DeepTEN} &
\textbf{GAP} &
\textbf{Histogram} &
\textbf{FAP} &
\textbf{Accuracy} \\ \hline
 & {Yes} & {Yes} & {No}& {No}&82.7 \\
 & {Yes} & {No}& {Yes} & {No}&81.9 \\
 & {Yes} & {No}& {No}& {Yes} &82.3 \\
 & {No}& {No}& {Yes} & {Yes} &52.8 \\
 & {No}& {Yes} & {Yes} & {No}&80.2 \\ 
 \multirow{-6}{*}{FMD} &
{No} &
{Yes} &
{No} &
{Yes} &77.4 \\ 
\hline
\end{tabular}}
\end{center}
\caption{Comparing ensemble of two out of four techniques at once}
\label{tab:2_by_4}
\end{table}
\begin{table}
\begin{center}
\begin{tabular}{|l|l|l|}
\hline
\textbf{Method} & \textbf{FMD} & \textbf{KTH} \\ 
\hline
{DeepTEN} & { 80.0} & {86.245}\\ 
\hline
GAP& {79.4} & {85.62} \\ 
\hline
HISTOGRAM & {72.9} & {86.027} \\ 
\hline
FAP& {33.8} & {60.52} \\ 
\hline
\end{tabular}
\end{center}
\caption{Mean accuracy of each technique applied individually on FMD and KTH}
\label{tab:1_by_4}
\end{table}
\begin{table*}[t]
\begin{center}
\scalebox{1.1}{%
\begin{tabular}{|l|l|l|l|l|l|l|}
\hline
\textbf{Method}& \textbf{FMD}& \textbf{KTH} & \textbf{DTD} & \textbf{MINC} & \textbf{GTOS} & \textbf{GTOS-M}  \\ \hline
% \multicolumn{1}{|c|}
DeepTEN & - & - & - & - & - & 76.12\\ \hline
DEPNet & - & - & - & - & - & 82.18\\ \hline
DSRNET &81.3 ± 0.8 & 81.8 ± 1.6 & 71.2 ± 0.7 &  - & 81.0 ± 2.1 & 83.65 ± 1.5\\ \hline
MAPNET &80.8 ± 1.0 & 80.9 ± 1.8 & 69.5 ± 0.8 & - & 80.3 ± 2.6 & 82.98 ± 1.6\\ \hline
LSCNET & 76.3 ± 0.1& - &  - & -  & -  & - \\ \hline
FENET & 82.3 ± 0.3& 86.6 ± 0.1 & 69.59 ± 0.04& 80.57 ± 0.10&83.10 ± 0.23&85.10 ± 0.36\\ \hline
HistNet & - & - & - & - & - & 79.75 ± 0.8\\ \hline
{CLASSNET} &82.5 ± 0.7 & 85.4 ± 1.1 & \textbf{71.5 ± 0.4} & 80.5 ± 0.6 & \textbf{84.3 ± 2.2} & \textbf{85.25 ± 1.3}\\ \hline
\textbf{Ours} & \textbf{83.10 ± 0.83}& \textbf{88.047 ± 3.45} & 70.208 ± 1.01 & \textbf{80.80 ± 0.44} & {83.38 ± 1.85} & 83.56 ± 0.0 \\ \hline
\end{tabular}}
\end{center}
\caption{Classification accuracies (\%) in the form of "mean±s.t.d." on the ResNet18 backbone. Best results are boldfaced}
\label{tab:SOTA}
\end{table*}
We conducted additional analysis on the ensemble methods. This was to study the TE technique performance when used alone, in combination of two techniques, and in combination of three techniques. The results for ensembling three techniques at the same time can be seen in Table \ref{tab:3_by_4}, for FMD and KTH dataset. Table \ref{tab:2_by_4} depicts the results for two techniques on FMD dataset. Table \ref{tab:1_by_4}, demonstrates how each technique uniquely performs on FMD.
\subsection{Effectiveness of Proposed Ensemble Technique}
Our proposed ensemble technique outperforms published research on models which have shown SOTA results on texture based classification. Our technique works on the principle of combining diverse techniques to produce a better result than any other technique used alone. Various combinations were researched and experimented on as part of this work as can be seen from Tables \ref{tab:3_by_4}, \ref{tab:2_by_4}, and \ref{tab:1_by_4}. It turns out that the combination of DeepTEN, FAP, and Histogram performs extremely well on all 6 benchmark datasets. We produce new SOTA results on KTH, FMD and MINC dataset. The more general way to look at this framework is to combine as many effective methods as possible for the classification task. However, When all the four methods were ensembled, we observe a dip in the accuracy in contrast to the combination of only three methods. This suggests that more investigation is needed in selecting the techniques to be used in the ensemble. The general rule is to combine techniques together which have different working principles. Suppose a Model $\mathcal{A}$ individually gives the correct prediction on a certain image while Model $\mathcal{B}$ individually gives a wrong prediction on the same image, when combined together in an ensemble, the network design changes and the model weights would result in either correct prediction or a wrong prediction. Thus the combination of methods should be properly investigated before ensembling is employed.

\subsection{More Analysis}
For the four techniques considered- FAP, GAP, DeepTEN, and HISTOGRAM, we tried experimenting with various possible combinations as per Tables \ref{tab:3_by_4}, \ref{tab:2_by_4}, and \ref{tab:1_by_4}. Total number of these experiments is 15 on single dataset FMD. As shown in Table \ref{tab:3_by_4} and \ref{tab:2_by_4} the highest accuracy obtained on FMD-related experiments is 83.10 by combining the three techniques DeepTEN, Histogram, and FAP. The peak accuracy for the combination of two methods- GAP and DeepTEN is 82.7, whereas, the peak accuracy for a single method (DeepTEN) is 80.0. We can observe that DeepTEN plays a major role in attaining the highest accuracy in every combination. In the three out of four methods combination experiment, if we remove DeepTEN the accuracy dropped to 80.7 which clearly demonstrates the importance of DeepTEN as a TE technique. Moreover, in the two out of four combinations, the second and third highest accuracy was obtained by combining DeepTEN with FAP (82.3) and DeepTEN with Histogram (81.9). 
\begin{figure}[t]
\begin{center}
\includegraphics[scale=0.2]{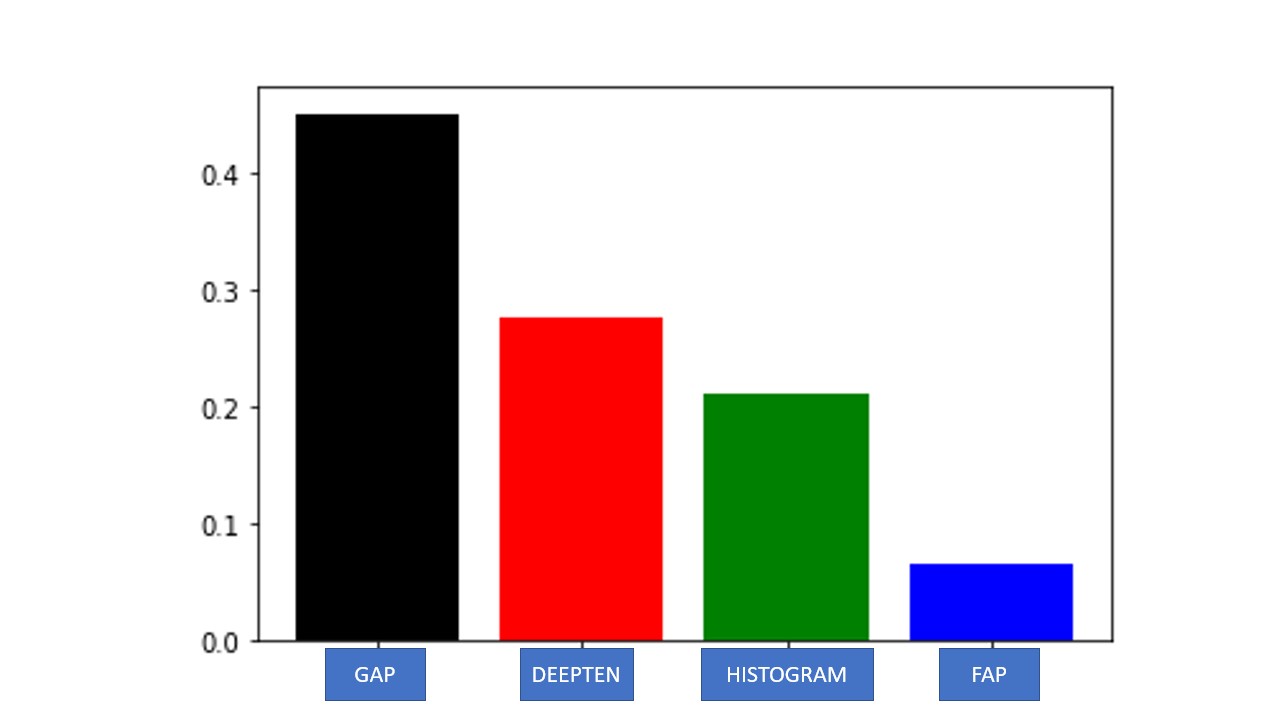}
\end{center}
\caption{Importance of techniques in deciding the accuracy}
\label{fig:feature_i}
\end{figure}

Another observation- when Histogram layer is removed from the combination of DeepTEN, Histogram and FAP the accuracy drops by merely 0.8, while, when FAP is removed the accuracy drops by 1.2 percent. Thus, it seems, FAP plays a major role when it is in the combination but not as an individual method because it is giving (worse) 33.8 accuracy on FMD when tested alone. GAP on the other hand has a neutral impact when combined with any of the other three, increasing with Histogram and DeepTEN separately, nonetheless, accuracy drops when it is combined with FAP. In the trio combination, it still shows a minimal increase in accuracy considering FAP is not in the trio. When GAP is removed, the accuracy reaches its highest value 83.10, on the contrary, when it’s introduced in the combination of all TE methods the accuracy drops by 1.1 percent. We attempted to find the feature importance of each of the four techniques used for classification task on FMD dataset. For this, we considered all 15 experiments on FMD dataset (all possible method combinations from 4 techniques). To achieve the feature importance values, we formulated a regression problem where we assumed accuracy values as regression values and encoded the presence of any technique in given experiment as 1 while absence as 0. We implemented Random Forest Regression algorithm on the formulated problem and calculated the feature importance on the trained model. We obtained the Figure \ref{fig:feature_i} by this procedure. It suggests, GAP has higher feature importance. In our case, it communicates that adding GAP tends to reduce the accuracy, on the contrary, removing GAP increases the accuracy. This is not always the case, for our model adding GAP can produce better results for datasets like GTOS \& GTOS-M. Furthermore, DeepTEN has the second highest feature importance value, which conveys that adding DeepTEN helps in increasing the accuracy while removing it reduces the accuracy values in context of our experimental setup. Histogram and FENet have the third and the least feature importance values respectively.

\section{Conclusion}
In summary, we propose a novel approach combining diverse TE techniques for improved learning compared to when these techniques are used individually. We develop a framework that incorporates two or more  texture-related techniques in a scalable way and demonstrate its effectiveness on benchmark texture datasets. The suggested framework not only delivers SOTA results but helps explain the impact of using dissimilar techniques in combination. We also produce better results on the larger datasets e.g. MINC, GTOS, and GTOS-M in less number of epochs i.e., 20 epochs as compared to 30 epochs in the cited papers covering various techniques. 

Our innovation offers an experimentally-proven combination of the existing texture extraction methods, efficiently, in terms of training steps and no. of parameters, eliminates the complexity of manual method selection and dependency on the training dataset. Our novelty is in research and presentation of a holistic architecture for extraction of diverse Texture patterns from images. With almost same number of parameters this approach produces SOTA results on 3 datasets in only 20 epochs as compared to 30 epochs in all the other cited methods. None of the previous research work covers the all listed benchmark texture datasets demonstrating results equally well across all.

In table 2 and 3 of the paper, we prove that using a random combination of the existing techniques does not produce SOTA results, therefore, the novel framework with its embedded specific texture extraction techniques proposed is non trivial.

We present our analysis on importance of GAP with respect to texture-based image datasets and discover that GAP does not exhibit significance for texture classification. In this paper, ensemble methods produce an improved results as compared to individual methods. Our ensembled architecture is simple and can easily be employed in standard CNNs as well as complex CNN architectures.
We combined three very different TE methods- DeepTEN, wherein feature extraction, dictionary learning, and encoding representation- all happen at the same time, FAP which focuses on discriminating textures based on the fractal dimension, and the histogram layer which captures the texture information directly from feature maps, via local histograms. Thus, it is up to the practitioners to try and use suitable TE method combinations to make a model perform well on other datasets.

We have considered 6 popular benchmark texture datasets to show the effectiveness of the proposed approach and achieved SOTA results on FMD, KTH, and MINC in this work. For the other 3 datasets our results are comparable with current SOTA results, achieving optimal results within 20 epochs of training. Our ensemble is scalable- other techniques can be added and existing ones can be removed from the existing architecture to experiment with other datasets. Our future work will entail investigating which TE methods to select while combining unique feature extractors based on the texture types of the given dataset with the goal of improving outcomes.

% \section{Acknowledgement}
% We would like to extend our deep gratitude to the team at Nvidia, Bengaluru Data centre for their support and encouragement throughout this body of work. Nvidia enabled us with world-class computing resources(a powerful DGX system with 8 x A100 32 GB GPUs) for running all the experiments.

{\small
\bibliographystyle{ieee_fullname}
\bibliography{main}

\begin{thebibliography}{10}\itemsep=-1pt

\bibitem{andrearczyk2016using}
Vincent Andrearczyk and Paul~F Whelan.
\newblock Using filter banks in convolutional neural networks for texture
  classification.
\newblock {\em Pattern Recognition Letters}, 84:63--69, 2016.

\bibitem{bell2015material}
Sean Bell, Paul Upchurch, Noah Snavely, and Kavita Bala.
\newblock Material recognition in the wild with the materials in context
  database.
\newblock In {\em Proceedings of the IEEE conference on computer vision and
  pattern recognition}, pages 3479--3487, 2015.

\bibitem{bors2001introduction}
Adrian~G Bors.
\newblock Introduction of the radial basis function (rbf) networks.
\newblock In {\em Online symposium for electronics engineers}, volume~1, pages
  1--7. DSP Algorithms San Jose, CA, USA, 2001.

\bibitem{bu2019deep}
Xingyuan Bu, Yuwei Wu, Zhi Gao, and Yunde Jia.
\newblock Deep convolutional network with locality and sparsity constraints for
  texture classification.
\newblock {\em Pattern Recognition}, 91:34--46, 2019.

\bibitem{caputo2005class}
Barbara Caputo, Eric Hayman, and P Mallikarjuna.
\newblock Class-specific material categorisation.
\newblock In {\em Tenth IEEE International Conference on Computer Vision
  (ICCV’05) Volume 1}, volume~2, page 1597—1604. IEEE, 2005.

\bibitem{chen2021deep}
Zhile Chen, Feng Li, Yuhui Quan, Yong Xu, and Hui Ji.
\newblock Deep texture recognition via exploiting cross-layer statistical
  self-similarity.
\newblock In {\em Proceedings of the IEEE/CVF Conference on Computer Vision and
  Pattern Recognition}, pages 5231--5240, 2021.

\bibitem{cimpoi2014describing}
Mircea Cimpoi, Subhransu Maji, Iasonas Kokkinos, Sammy Mohamed, and Andrea
  Vedaldi.
\newblock Describing textures in the wild.
\newblock In {\em Proceedings of the IEEE conference on computer vision and
  pattern recognition}, page 3606—3613, 2014.

\bibitem{cimpoi2015deep}
Mircea Cimpoi, Subhransu Maji, and Andrea Vedaldi.
\newblock Deep filter banks for texture recognition and segmentation.
\newblock In {\em Proceedings of the IEEE conference on computer vision and
  pattern recognition}, pages 3828--3836, 2015.

\bibitem{dai2017fason}
Xiyang Dai, Joe Yue-Hei~Ng, and Larry~S Davis.
\newblock Fason: First and second order information fusion network for texture
  recognition.
\newblock In {\em Proceedings of the IEEE Conference on Computer Vision and
  Pattern Recognition}, pages 7352--7360, 2017.

\bibitem{freeman1997learning}
William~T Freeman and Joshua~B Tenenbaum.
\newblock Learning bilinear models for two-factor problems in vision.
\newblock In {\em Proceedings of IEEE Computer Society Conference on Computer
  Vision and Pattern Recognition}, pages 554--560. IEEE, 1997.

\bibitem{fujieda2017wavelet}
Shin Fujieda, Kohei Takayama, and Toshiya Hachisuka.
\newblock Wavelet convolutional neural networks for texture classification.
\newblock {\em arXiv preprint arXiv:1707.07394}, 2017.

\bibitem{he2016deep}
Kaiming He, Xiangyu Zhang, Shaoqing Ren, and Jian Sun.
\newblock Deep residual learning for image recognition.
\newblock In {\em Proceedings of the IEEE conference on computer vision and
  pattern recognition}, pages 770--778, 2016.

\bibitem{julesz1981textons}
Bela Julesz.
\newblock Textons, the elements of texture perception, and their interactions.
\newblock {\em Nature}, 290(5802):91--97, 1981.

\bibitem{lin2013network}
Min Lin, Qiang Chen, and Shuicheng Yan.
\newblock Network in network.
\newblock {\em arXiv preprint arXiv:1312.4400}, 2013.

\bibitem{lin2016visualizing}
Tsung-Yu Lin and Subhransu Maji.
\newblock Visualizing and understanding deep texture representations.
\newblock In {\em Proceedings of the IEEE conference on computer vision and
  pattern recognition}, pages 2791--2799, 2016.

\bibitem{liu2019bow}
Li Liu, Jie Chen, Paul Fieguth, Guoying Zhao, Rama Chellappa, and Matti
  Pietik{\"a}inen.
\newblock From bow to cnn: Two decades of texture representation for texture
  classification.
\newblock {\em International Journal of Computer Vision}, 127(1):74--109, 2019.

\bibitem{liu2022deep}
Xiu Liu and Chris Aldrich.
\newblock Deep learning approaches to image texture analysis in material
  processing.
\newblock {\em Metals}, 12(2):355, 2022.

\bibitem{loshchilov2016sgdr}
Ilya Loshchilov and Frank Hutter.
\newblock Sgdr: Stochastic gradient descent with warm restarts.
\newblock {\em arXiv preprint arXiv:1608.03983}, 2016.

\bibitem{lowe2004distinctive}
David~G Lowe.
\newblock Distinctive image features from scale-invariant keypoints.
\newblock {\em International journal of computer vision}, 60(2):91--110, 2004.

\bibitem{mandelbrot1967long}
Benoit Mandelbrot.
\newblock How long is the coast of britain? statistical self-similarity and
  fractional dimension.
\newblock {\em science}, 156(3775):636--638, 1967.

\bibitem{pandey2020incorporating}
Vijay Pandey and Shashi~Bhushan Jha.
\newblock Incorporating image gradients as secondary input associated with
  input image to improve the performance of the cnn model.
\newblock {\em arXiv preprint arXiv:2006.04570}, 2020.

\bibitem{peeples2021histogram}
Joshua Peeples, Weihuang Xu, and Alina Zare.
\newblock Histogram layers for texture analysis.
\newblock {\em IEEE Transactions on Artificial Intelligence}, 2021.

\bibitem{petersson2016multi}
Henrik Petersson and David Gustafsson.
\newblock Multi-spectral texture analysis for ied detection.
\newblock In {\em Electro-Optical Remote Sensing X}, volume 9988, pages
  206--219. SPIE, 2016.

\bibitem{shanmugavadivu2012fractal}
P Shanmugavadivu and V Sivakumar.
\newblock Fractal dimension based texture analysis of digital images.
\newblock {\em Procedia Engineering}, 38:2981--2986, 2012.

\bibitem{sharan2009material}
Lavanya Sharan, Ruth Rosenholtz, and Edward Adelson.
\newblock Material perception: What can you see in a brief glance?
\newblock {\em Journal of Vision}, 9(8):784—784, 2009.

\bibitem{silven2000applying}
O Silven.
\newblock Applying texture analysis to industrial inspection.
\newblock In {\em Texture Analysis in Machine Vision}, pages 207--217. World
  Scientific, 2000.

\bibitem{tuceryan1993texture}
Mihran Tuceryan and Anil~K Jain.
\newblock Texture analysis.
\newblock {\em Handbook of pattern recognition and computer vision}, pages
  235--276, 1993.

\bibitem{xu2021encoding}
Yong Xu, Feng Li, Zhile Chen, Jinxiu Liang, and Yuhui Quan.
\newblock Encoding spatial distribution of convolutional features for texture
  representation.
\newblock {\em Advances in Neural Information Processing Systems}, 34, 2021.

\bibitem{xue2018deep}
Jia Xue, Hang Zhang, and Kristin Dana.
\newblock Deep texture manifold for ground terrain recognition.
\newblock In {\em Proceedings of the IEEE Conference on Computer Vision and
  Pattern Recognition}, pages 558--567, 2018.

\bibitem{zhai2020deep}
Wei Zhai, Yang Cao, Zheng-Jun Zha, HaiYong Xie, and Feng Wu.
\newblock Deep structure-revealed network for texture recognition.
\newblock In {\em Proceedings of the IEEE/CVF Conference on Computer Vision and
  Pattern Recognition}, pages 11010--11019, 2020.

\bibitem{zhai2019deep}
Wei Zhai, Yang Cao, Jing Zhang, and Zheng-Jun Zha.
\newblock Deep multiple-attribute-perceived network for real-world texture
  recognition.
\newblock In {\em Proceedings of the IEEE/CVF International Conference on
  Computer Vision}, pages 3613--3622, 2019.

\bibitem{zhang2017deep}
Hang Zhang, Jia Xue, and Kristin Dana.
\newblock Deep ten: Texture encoding network.
\newblock In {\em Proceedings of the IEEE conference on computer vision and
  pattern recognition}, pages 708--717, 2017.

\bibitem{zhang2019learning}
Jiahui Zhang, Dawei Sun, Zixin Luo, Anbang Yao, Lei Zhou, Tianwei Shen, Yurong
  Chen, Long Quan, and Hongen Liao.
\newblock Learning two-view correspondences and geometry using order-aware
  network.
\newblock In {\em Proceedings of the IEEE/CVF international conference on
  computer vision}, pages 5845--5854, 2019.

\bibitem{zu2015real}
Zhaozi Zu, Yingtuan Hou, Dixiao Cui, and Jianru Xue.
\newblock Real-time road detection with image texture analysis-based vanishing
  point estimation.
\newblock In {\em 2015 IEEE International Conference on Progress in Informatics
  and Computing (PIC)}, pages 454--457. IEEE, 2015.

\end{thebibliography}
}
\end{document}